# Aeolus Ocean - A simulation environment for the autonomous COLREG-compliant navigation of Unmanned Surface Vehicles using Deep Reinforcement Learning and Maritime Object Detection

Andrew Alexander Vekinis
Computational Intelligence Laboratory
Institute of Informatics & Telecommunications,
National Center for Scientific Research "Demokritos", Athens, Greece

Stavros Perantonis
Computational Intelligence Laboratory
Institute of Informatics & Telecommunications,
National Center for Scientific Research "Demokritos", Athens, Greece

***Abstract*** – Heading towards navigational autonomy in unmanned surface vehicles (USVs) in the maritime sector can fundamentally lead towards safer waters as well as reduced operating costs, while also providing a range of exciting new capabilities for oceanic research, exploration and monitoring. However, achieving such a goal is challenging. USV control systems must, safely and reliably, be able to adhere to the international regulations for preventing collisions at sea (COLREGs) in encounters with other vessels as they navigate to a given waypoint while being affected by realistic weather conditions, either during the day or at night. To deal with the multitude of possible scenarios, it is critical to have a virtual environment that is able to replicate the realistic operating conditions USVs will encounter, before they can be implemented in the real world. Such 'digital twins' form the foundations upon which Deep Reinforcement Learning (DRL) and Computer Vision (CV) algorithms can be used to develop and guide USV control systems. In this paper we describe the novel development of a COLREG-compliant DRL-based collision avoidant navigational system with CV-based awareness in a realistic ocean simulation environment. The performance of the trained autonomous Agents resulting from this approach is evaluated in several successful navigations to set waypoints in both open sea and coastal encounters with other vessels. A binary executable version of the simulator with trained agents is available at: https://github.com/aavek/Aeolus-Ocean .

***Keywords*** – Deep Reinforcement Learning, Collision Avoidance, COLREGs, Unmanned Surface Vehicle, Maritime Autonomous Surface Ships (MASS), Digital Twin, Unity Engine, Object Detection, Computer Graphics, Physically-Based Modeling

## 1. Introduction

In recent years, there has been a significant upsurge in industrial research projects that have aimed to bring autonomy into the maritime commercial, research and defense sectors using Machine Learning (ML). Projects such as Mayflower[1], L3HARRIS[2], Cetus[3] and MAXCMAS[4], show that there is considerable interest in the investment of autonomous ship technologies, and the EU has demonstrated through funding programmes that it wants to incentivize the transition towards maritime autonomy [5].

The main hurdle behind the true adoption of MASS technologies however is regulatory [6]. Essentially, autonomy needs to be safe, secure and have a minimal impact to the environment. One the primary ways to ensure this is with the development of reliable collision avoidant control systems that are able to minimize risk during real-weather navigation, especially in encounters with other vessels near coastland or in ports. This of course requires the agent to be continuously aware of nearby vessels, accurately knowing their positions, which in itself is a challenge as it plans its path towards a given waypoint. This is what is known as the mission planning problem [7].

The topic of collision avoidance (COLAV) has been studied extensively; either using classical (non-ML) or ML approaches in virtual environments, in compliance with COLREGs which are the international regulations for preventing collisions at sea [8]. Classical approaches [9-14] implement COLAV systems using analytical approaches such as velocity obstacles, model predictive control and potential fields, typically in tandem with path following models. ML approaches predominantly use DRL methods [7] that learn control policies via feedback from designed reward functions. Many approaches in the literature however [15-16] are not able to satisfy all the following prerequisites for realism:

- Full (6 D.O.F.) motion dynamics of vessels that include pitch, roll and heave motions.
- Ability to deal with complex environments that have multiple dynamic and static obstacles while complying with COLREGs.
- Weather and Day/Night visibility effects on the awareness of other vessels.
- Wave and wind force impacts at different sea-states as well as underwater terrain hazards.

Some approaches are also limited by high computational costs and cannot scale to include enough scenarios required for generalization [7]. The current work aims to address each one of these points. We therefore propose, to our knowledge, a novel approach for the mission planning problem where we demonstrate the COLREG-compliant navigation of a USV towards waypoints using DRL while being aware of nearby obstacles using sensor systems that include CV algorithms, in a fully three-dimensional realistic synthetic ocean environment. Thus, the contributions of this paper are:

- A fully controllable realistic ocean environment complete with weather systems, adjustable time-of-day, a physics model that handles vessel-water interactions and propulsion, and a procedural terrain system.
- Demonstrating the viability of DRL in COLREG-compliant COLAV maneuvering towards a waypoint for unseen coastlines and vessel encounters.
- Implementing novel reward functions that are able to deal with the above requirements in the realistic ocean environment.
- Sensor fusion techniques that incorporate CV algorithms such as object detection for accurate vessel tracking in various operating conditions.

The last contribution could also aid in real-world situations where Automatic Identifications Systems (AIS) have been switched off.

The dimensions of the motorboat USV model chosen for this work [17] are approximately similar to [1-4] with an axial length of ~8m, weighing a simulated 3.5 tons.

Specifically, the DRL control policies for the USV will be trained using Proximal Policy Optimization (PPO) [18]. This is due to its superior performance in dealing with continuous COLAV actions [7] as it navigates the state-space towards globally defined waypoints. The observations of the environment will come from the sensor fusion of a simulated simplified radar system, a simplified sonar system, and the YOLO-X object detection algorithm [19] to provide accurate and up-to-date positions of nearby vessels and land proximity. To ensure sufficient object detection accuracy, the pre-trained weights for YOLO-X are fine-tuned on a mixed real/synthetic dataset with a curriculum learning strategy inspired by [20]. Optimal continuous control actions will be trained using a combination of reward functions that deal with waypoint navigation, dynamic COLAV between other vessels and static COLAV for land.

We evaluate the trained agent's performance on various vessel encounters (head-on, crossing, randomized) in the proximity of procedurally generated islands in different weather conditions. Finally we discuss the results and propose future directions to address ways in which the current approach could be improved.

## 2. Background

Each aspect of the simulator is now described, focusing firstly on the necessary mathematical background used to create the simulation environment. Then, the maritime navigation rules for COLREG-compliance are laid out. Finally, the fundamentals of both the DRL and CV techniques used in this paper are explained.

### I. Ocean Simulation Environment

We make use of the techniques employed by Tessendorf [21] and consequently Horvath [22] and Gamper [23] to model our ocean surface waves and the interested reader is highly encouraged to peruse these sources.

In summary, we use a wave spectral density function (or wave spectrum, $S$) and random numbers drawn from a Gaussian distribution to first create the initial wave state which is in a 2D spectral domain. As time increases, the dispersion relationship [21] is used to propagate these waves forward with a specific wavelength and phase velocity. Inverse Fast Fourier Transforms are then used to reconstruct the spatial form of the waves as 2D height maps, with the capability to control horizontal spatial warping for choppy waves and whitecap regions.

The ocean surface wave calculation procedure is now described. From Linear Wave Theory [24-26], we can expand to consider an infinite number of 2D waves, of different wavelengths and directions, summed up in three-dimensional space to produce a vertical displacement $\eta$ at each position $\boldsymbol{X} = (x, y)$ on the horizontal axis of an ocean surface, as defined by equation (1):

$$\eta(\boldsymbol{X}, t) = \sum_{i=0}^{\infty}[A_i \cos(\boldsymbol{k}_i^\mathrm{T}\boldsymbol{X} - \omega_i t + \varphi_i)] \quad (1)$$

Where subscript $i$ is the index of each wave, wave vector $\boldsymbol{k}_i = \left(k_{x_i}, k_{y_i}\right) = (k_i \cos\theta, k_i \sin\theta)$ is comprised of $k_i = 2\pi/\lambda_i$ with $\lambda_i$ as the wavelength of each wave and $\theta_i = arctan(k_i \sin\theta / k_i \cos\theta)$ being the direction of the wave from the x-axis, $A_i$ is the notional wave amplitude, $\omega_i$ and $t$ being the angular frequency and time elapsed respectively and $\varphi_i$ being the wave phase shift.

The wave number $k_i$ and the angular frequency $\omega_i$ are related via the dispersion relationship via equation (2) from [22], at a finite-depth and considering dynamics affected by a surface tension parameter, $\sigma \approx 0.074$. Here, $g$ is the acceleration of gravity and $D$ is the uniform sea depth with sea water density $\rho_{sea}$:

$$\omega_i = \sqrt{(gk_i + \frac{\sigma}{\rho_{sea}}k_i^3)tanh(k_i D)} \quad (2)$$

To make equation (1) useable in practice, we follow the process in [22] where in the spectral domain we can discretize the angular frequencies $\omega_i$ and wave directions $\theta_i$ in intervals of $\Delta\omega_i$ and $\Delta\theta_i$ with amplitudes $A_i$ at every $X$ so that:

$$\sum_{\Delta\omega_i} \sum_{\Delta\theta_i} \left[\frac{A_i^2}{2}\right] = \int_{-\pi}^{\pi}\int_0^{\infty} S(\omega,\theta)d\omega d\theta \quad (3)$$

Where for a single wave this would become:

$$\frac{A_i^2}{2} = \int_{\theta_i}^{\theta_i+\Delta\theta_i}\int_{\omega_i}^{\omega_i+\Delta\omega_i} S(\omega,\theta)d\omega d\theta \approx S(\omega_i,\theta_i)\Delta\theta_i\Delta\omega_i \quad (4)$$

Thus, from (4), it can be deduced that the vertical displacement of the ocean surface at every point $X$ for every discretized angular frequency $\omega_i$ and wave direction $\theta_i$ is related to a directional wave spectrum $S(\omega_i,\theta_i)$. This can be expanded into two terms, $S(\omega)$, the non-directional component of the wave spectrum that contains the frequency profile of the waves and $D(\omega,\theta)$, the directional spreading function, shown in equation (5):

$$S(\omega_i,\theta_i) = S(\omega)D(\omega,\theta) \quad (5)$$

**Non-Directional Wave Spectra, $S(\omega)$**

In our ocean model we implement two non-directional wave spectra which can be chosen at will during the simulation. These are the Pierson-Moskowitz (PM) Spectrum [27], the JONSWAP spectrum [28] and the Texel Marsen Arsloe (TMA) Spectrum [29] which is essentially is the product of the JONSWAP spectrum and the Kitaigorodskii Depth Attenuation Function [30]. These will now be described.

The PM spectrum is for fully-developed seas where the wind-generated waves are the largest they can be from for a given wind-speed at 10 meters above the surface, $U_{10}$. Essentially this spectrum assumes all wind energy has been extracted to move the wave before this energy is dissipated. It is defined by:

$$S_{PM}(\omega) = \frac{ag^2}{\omega^5}e^{\left[-\frac{5}{4}\left(\frac{\omega_p}{\omega}\right)^4\right]} \quad (6)$$

Where: $a = 8.1 \times 10^{-3}$

$$\omega_p \approx \frac{0.855g}{U_{10}}$$

The parameter $a$ is an empirical constant, while $\omega_p$ is the peak angular frequency; the frequency at which the waves have the most average energy extracted from the wind.

The JONSWAP spectrum is for developing seas, where the waves still have the potential to grow and evolve as a constant wind blows over them over a distance from shore called *fetch*, $F$. It should be noted that even for large $F$, $S_{\text{JONSWAP}}(\omega) \neq S_{PM}(\omega)$. The JONSWAP spectrum is defined by, following [23]:

$$S_{\text{JONSWAP}}(\omega) = \frac{ag^2}{\omega^5}\gamma^r e^{\left[-\frac{5}{4}\left(\frac{\omega_p}{\omega}\right)^4\right]} \quad (7)$$

Where:
$$a = 0.076\chi^{-0.22}$$
$$\chi = \frac{gF}{U_{10}^2}, \quad \gamma = 3.3$$
$$r = e^{\left[-\frac{(\omega-\omega_p)^2}{2\sigma^2\omega_p^2}\right]}$$
$$\omega_p = 2\pi\frac{g\nu}{U_{10}}, \quad \nu = 3.5\,\chi^{-0.33}$$
$$\sigma = \begin{cases}0.07, & \omega \leq \omega_p \\ 0.09, & \omega > \omega_p\end{cases}$$

The peak angular frequency $\omega_p$ is extracted from the non-dimensional peak wave frequency $\nu$ which is dependent on the non-dimensional fetch, $\chi$. The term $\gamma^r$ is the peak enhancement factor and $\sigma$ is the peak width parameter.

The TMA Spectrum effectively enhances the JONSWAP spectrum via the Kitaigorodskii Depth Attenuation Function $\Phi(\omega,D)$, defined by:

$$\Phi(\omega,D) = \begin{cases}\frac{\omega_h^2}{2}, & \omega_h \leq 1 \\ 1 - \frac{1}{2}(2-\omega_h)^2, & \omega_h > 1\end{cases} \quad (8)$$

Where: $\omega_h = \omega\sqrt{\frac{D}{g}}$

Therefore the TMA Spectrum is simply expressed as:

$$S_{\text{TMA}}(\omega) = S_{\text{JONSWAP}}(\omega) \cdot \Phi(\omega,D) \quad (9)$$

**Directional Spreading Function, $D(\omega,\theta)$**

Our ocean model uses the Hasselmann [28] directional spreading factor, denoted by $s$. $D(\omega,\theta)$ is specified by:

$$D(\omega,\theta) = \frac{2^{2s-1}}{\pi}\frac{\Gamma^2(s+1)}{\Gamma(2s+1)}\left|\cos\left(\frac{\theta_i-\theta_p}{2}\right)\right|^{2s} \quad (10)$$

Where $\Gamma$ is the gamma function, which we estimate using Ramanujan's approximation formula [31]:

$$\Gamma(s+1) = \sqrt{\pi}\left(\frac{s}{e}\right)^s\left(8s^3+4s^2+s+\frac{1}{30}\right)^{1/6} \quad (11)$$

The term $\theta_i$ is the direction of the wave described in (1), while $\theta_p$ is the direction of the wind in radians from the x-axis. The Hasselmann Spreading Factor, $s$ is:

$$s = \begin{cases} 9.77 \left(\frac{\omega}{\omega_p}\right)^{\mu} & , \omega > \omega_p \\ 6.97 \left(\frac{\omega}{\omega_p}\right)^{4.06} & , \omega \leq \omega_p \end{cases} \quad (12)$$

Where: $\mu = -2.33 - 1.45 \left(\frac{U_{10}\omega_p}{g} - 1.17\right)$

To incorporate the effects of ocean swell if required, following [22], we can add to the Hasselmann Spreading Factor, $s$ in (12) an elongation term $s_\xi$, controlled by a wave alignment factor, $\xi \in [0,1]$ as shown in (13):

$$s_\xi = 16\tanh\left(\frac{\omega}{\omega_p}\right)\xi^2 \quad (13)$$

**Ocean Surface Waves Synthesis**

The final step in the ocean surface wave calculation procedure is to transform the directional wave spectra, $S(\omega, \theta)$ into two-dimensional wave number spectra with wave vectors, $S(\mathbf{k})$ before using inverse FFTs to transform to the spatial domain. Following [22], this is done through the change of variables theorem which will not be discussed here. Equation (14) shows the transformation:

$$S(\mathbf{k}) = S(\omega, \theta)\frac{d\omega}{dk}\frac{1}{k} = S(\omega, \theta)\left|\frac{df(k)}{dk}\frac{1}{k}\right| \quad (14)$$

Where $k = \sqrt{k_x^2 + k_y^2}$. Extracting the partial derivatives via chain-rule of the dispersion relation $\omega = f(k)$ of (2) makes (14) possible because we can approximate by:

$$\Delta\omega\Delta\theta \approx \left|\frac{df(k)}{dk}\frac{1}{k}\right|\Delta k_x \Delta k_y \quad (15)$$

Thus, the spatial representation of the wave spectra, with amplitudes $A$, is obtained by rearranging (4):

$$A_{(\mathbf{k},\ \Delta k_x,\ \Delta k_y)} = R\sqrt{2S(\mathbf{k})\Delta k_x \Delta k_y} \quad (16)$$

Where $R \sim N(0,1)$ across every horizontal position $\mathbf{X}$.

To optimize the visual fidelity of the ocean we also implement a multi-resolution approach [23] by stacking multiple outputs from (16) with similar wind directions at different scales, to form compound wave spectra and thus accommodate for finer details.

## II. Collision Regulations (COLREGs)

In maritime environments there exist internationally agreed regulations for preventing collisions at sea [8], written as a set of navigational rules that must be followed when vessels encounter each other. We will focus specifically, but briefly, on the rules that are involved with COLAV maneuvering, as they are listed below:

**NAVIGATION RULES**
**PART B - STEERING AND SAILING RULES**

*I - Conduct of Vessels in Any Condition of Visibility*

**Rule 6 - Safe Speed**
*Every vessel shall at all times proceed at a safe speed so that she can take proper and effective action to avoid collision and be stopped within a distance appropriate to the prevailing circumstances and conditions.*

**Rule 8 - Action to Avoid Collision**
*(a) Any action taken to avoid collision shall ... if the circumstances of the case admit, be positive, made in ample time and with due regard to the observance of good seamanship.*

*(b) Any alteration of course and/or speed to avoid collision shall, if the circumstances of the case admit, be large enough to be readily apparent to another vessel observing visually or by radar; a succession of small alterations of course and/or speed should be avoided.*

*(c) If there is sufficient sea room, alteration of course alone may be the most effective action to avoid a close-quarters situation provided that it is made in good time, is substantial and does not result in another close-quarters situation.*

*(d) Action taken to avoid collision with another vessel shall be such as to result in passing at a safe distance. The effectiveness of the action shall be carefully checked until the other vessel is finally past and clear.*

*(e) If necessary to avoid collision or allow more time to assess the situation, a vessel shall slacken her speed or take all way off by stopping or reversing her means of propulsion.*

*II - Conduct of Vessels in Sight of One Another*

**Rule 13 – Overtaking**
*(b) A vessel shall be deemed to be overtaking when coming up with a another vessel from a direction more than 22.5 degrees abaft her beam, that is, in such a position with reference to the vessel she is overtaking, that at night she would be able to see only the sternlight of that vessel but neither of her sidelights.*

*(d) Any subsequent alteration of the bearing between the two vessels shall not make the overtaking vessel a crossing vessel within the meaning of these Rules or relieve her of the duty of keeping clear of the overtaken vessel until she is finally past and clear.*

**Rule 14 - Head-on Situation**
*(a) Unless otherwise agreed, when two power-driven vessels are meeting on reciprocal or nearly reciprocal courses so as to involve risk of collision each shall alter her course to starboard so that each shall pass on the port side of the other.*

*(b) Such a situation shall be deemed to exist when a vessel sees the other ahead or nearly ahead and by night she could see the masthead lights of the other in a line or nearly in a line and/or both sidelights and by day she observes the corresponding aspect of the other vessel.*

*(c) When a vessel is in any doubt as to whether such a situation exists she shall assume that it does exist and act accordingly.*

**Rule 15 - Crossing Situation**
*When two power-driven vessels are crossing so as to involve risk of collision, the vessel which has the other on her own starboard side shall keep out of the way and shall, if the circumstances of the case admit, avoid crossing ahead of the other vessel.*

**Rule 16 - Action by Give-way Vessel**

*Every vessel which is directed to keep out of the way of another vessel shall, so far as possible, take early and substantial action to keep well clear.*

*III - Conduct of Vessels in Restricted Visibility*

**Rule 19 - Conduct of Vessels in Restricted Visibility**

*(b) Every vessel shall proceed at a safe speed adapted to the prevailing circumstances and conditions of restricted visibility. A power-driven vessel shall have her engines ready for immediate maneuver.*

*(d) A vessel which detects by radar alone the presence of another vessel shall determine if a close-quarters situation is developing and/or risk of collision exists. If so, she shall take avoiding action in ample time, provided that when such action consists of an alteration in course, so far as possible the following shall be avoided:*
*(i) An alteration of course to port for a vessel forward of the beam, other than for a vessel being overtaken;*
*(ii) An alteration of course toward a vessel abeam or abaft the beam.*

We focus on the compliance of the DRL Agent (EGO vessel) with each of the aforementioned Rules, especially in the context of changes in visibility. Figure 1 below provides a graphical summary of the basic encounters and the rules that need to be followed:

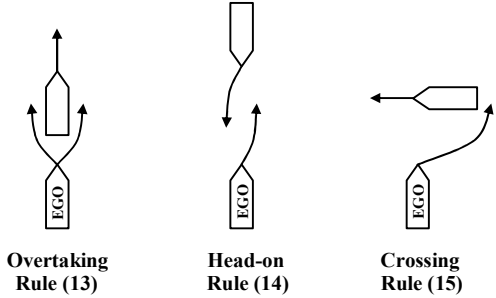

**Overtaking Rule (13)**    **Head-on Rule (14)**    **Crossing Rule (15)**

**Figure 1: Recommended actions to be taken by the EGO-vessel in basic encounters for COLREG rules 13, 14, 15.**

## III. Deep Reinforcement Learning

The purpose of Reinforcement Learning (RL) is to develop optimal decision-making policies for Agents as they interact with their Environment. These policies, which guide the actions Agents take towards the attainment of specific goals, are learnt by trial and error through reward feedback. A policy that can maximize all specified rewards over time is therefore the task at hand. It should be noted that the rewards themselves are manually designed to encourage the development of a required policy, and are described in more detail in the Methodology section of this paper.

In the context of control systems for autonomous vessels, many approaches proposed in the literature [7] have used model-free RL methods such as Q-Learning [32] and Policy Optimization with Gradient Methods. Examples of the former are Deep Q-Networks (DQN) [33], while for the latter these have included Asynchronous Advantage Actor-Critic (A3C) [34], Deep Deterministic Policy Gradient (DDPG) [35] and Proximal Policy Optimization (PPO) [18].

In this paper we elect to use the Proximal Policy Optimization (PPO) algorithm with continuous action spaces as it has been shown to exhibit superior performance in the mission planning problem we are trying to solve when compared to other algorithms [15][36]. The following fundamentals will now be explained in the context of PPO.

We first formally express the dynamics between the Agent and the Environment as a 6-tuple Markov Decision Process (MDP), $\langle S, A, P, R, \rho_0, \gamma \rangle$ where:

- $S$ is a set of all finite possible environmental states,
- $A$ is a set of all finite possible actions to be taken,
- $P$ is the function for the probability of transitioning into the next time-step state $s_{t+1}$, given the current state $s_t$ and action $a_t$, i.e. $P(s_{t+1} \mid s_t, a_t)$,
- $R$ is the reward function that yields a scalar reward $r_t$, for an action $a_t$, taken at state $s_t$, i.e. $r_t = R(s_t, a_t)$,
- $\rho_0$ is the state-space distribution at the start,
- $\gamma$ is a discount factor for balancing between distant and near future rewards, where $\gamma \in [0,1]$.

Actions $a_t$ are informed by a parameterized policy $\pi_\theta(s_t)$, where the parameters $\theta$ can be the weights and biases of a deep neural network which can be adjusted to some optimal form $\theta^*$ using optimization algorithms. In the case of continuous action spaces, actions $a_t \sim \pi_\theta(s_t)$ are stochastically sampled from policies drawn from a multivariate Gaussian distribution, tuned by $\theta$. In essence, the reward signals $r_t$ extracted as a result from these actions should alter $\theta$ in such a way so that the probability of getting higher rewards is increased. It should be noted that the part of the network that executes a policy is called an 'Actor'. The resulting action will lead to an output value (in state-space) $V_\phi$ with parameters $\phi$. This is the 'Critic' part of the network.

A collection of time-steps can form a finite trajectory, or episode, $\tau = (s_0, a_0, s_1, a_1, \ldots s_t, a_t, \ldots s_T, a_T)$ which is defined as a sequence of events in the environment, each with a state-action pair at time-step, $t$, until the end-time of the episode, $T$. By trying out different actions, our policy $\pi_\theta$ will generate a probability for a sequence of state-actions $P(\tau|\pi_\theta)$ that have a cumulative reward $R(\tau)$ from a sequence of rewards $r_t$ discounted by $\gamma$. These are defined by equations (17) and (18):

$$P(\tau|\pi_\theta) = \rho_0(s_0) \prod_{t=0}^{T-1} [P(s_{t+1} \mid s_t, a_t) \cdot \pi(a_t, s_t)]$$

(17)

$$R(\tau) = \sum_{t=0}^{\infty} \gamma^t r_t \tag{18}$$

We can now express the RL problem as an optimization problem as shown in equation (19):

$$J(\pi_\theta) = \int_\tau P(\tau|\pi_\theta) \cdot R(\tau) = E_{\tau \sim \pi_\theta} R(\tau) \tag{19}$$

Where we try to maximize the expected reward by choosing an optimal policy, $\pi_{\theta^*}$ as shown in (20):

$$\pi_{\theta^*} = \underset{\pi_\theta}{\mathrm{argmax}}[J(\pi_\theta)] \tag{20}$$

**Policy Gradients**

The optimal policy $\pi_{\theta^*}$ can be obtained by updating the parameters $\theta$ using stochastic gradient ascent with learning rate, $\lambda \in [0,1]$, shown in (21):

$$\theta \longleftarrow \theta + \lambda \nabla_\theta J(\pi_\theta) \tag{21}$$

Here, $\nabla_\theta J(\pi_\theta)$ is known as the policy gradient. From the derivations in [37] and by introducing an Advantage function for a current policy, $A^{\pi_\theta}(s_t, a_t)$, the policy gradient can be expressed in (22) as:

$$\nabla_\theta J(\pi_\theta) = E_{\tau \sim \pi_\theta} \left[ \sum_{t=0}^{T} \nabla_\theta \log(\pi_\theta(a_t|s_t)) \cdot A^{\pi_\theta}(s_t, a_t) \right] \tag{22}$$

Where:

$$A^{\pi_\theta}(s_t, a_t) = Q^\pi(s_t, a_t) - V^\pi(s_t) \tag{23}$$

The Advantage function in (23) compares how beneficial a specific action $a_t$ at current state $s_t$, sampled from a policy $\pi_\theta$, is to other actions on average in that state. The terms $Q^\pi(s_t, a_t)$ and $V^\pi(s_t)$ can give us the value of a state are:

$$Q^\pi(s_t, a_t) = E_{\tau \sim \pi_\theta}[R(\tau)|s_0 = s, a_0 = a] \tag{24}$$

$$V^\pi(s_t) = E_{\tau \sim \pi_\theta}[R(\tau)|s_0 = s] \tag{25}$$

To make (22) tractable, we can approximate the policy gradient over $N$ episodes using:

$$\nabla_\theta J(\pi_\theta) = \frac{1}{N} \sum_{n=1}^{N} \sum_{t=0}^{T} \nabla_\theta \log(\pi_\theta(a_t^n|s_t^n)) \cdot (A^{\pi_\theta}(s_t, a_t))^n \tag{26}$$

Values for (24), $V^\pi \to V_\phi$, and (25) can be approximated via neural networks using the Generalized Advantage Estimation (GAE) method following [38].

**Proximal Policy Optimization**

In the PPO algorithm, equation (26), and thus (21) are not used directly. The reason for this is that if an update to a policy is too large, the policy can become unstable, leading to a performance collapse. On the other hand, we do want updates to a policy to as large as possible to speed up convergence to an optimal level. To do this we measure how much a current policy changed compared the older policy (at epoch $K$) in the previous update and constrain the allowable policy change via a hyperparameter $\epsilon$ (typically 0.2). This leads to:

$$\theta_{K+1} = \underset{\theta}{\mathrm{argmax}} \, E_{s,a \sim \pi_{\theta_K}}(J^{CLIP}) \tag{27}$$

Here, $J^{CLIP}$ is a function of $s, a, \theta_K$ and $\theta$, also known as the clipped surrogate objective function, expressed in equation (28) from [17] as:

$$J^{CLIP} = \min\left(\frac{\pi_\theta(a|s)}{\pi_{\theta_k}(a|s)} A^{\pi_{\theta_K}}(s,a), g(\epsilon, A^{\pi_{\theta_K}}(s,a))\right) \tag{28}$$

Where:

$$g = \begin{cases} (1+\epsilon)A^{\pi_{\theta_K}} &, A^{\pi_{\theta_K}} \geq 0 \\ (1-\epsilon)A^{\pi_{\theta_K}} &, A^{\pi_{\theta_K}} < 0 \end{cases} \tag{29}$$

A final point to make is that (18) is also altered, so that instead of summing all rewards that were ever obtained, rewards are instead extracted *after* actions have been completed. Thus, we have the 'reward-to-go' $\hat{R}_t$ which is what will be used to compare against $V_\phi^\pi$ in the critic network, shown in (30):

$$\hat{R}_t = \sum_{t'-t}^{T} R(s_{t'}, a_{t'}, s_{t'+1}) \tag{30}$$

Therefore, the process is summarized in Algorithm 1, considering multiple single-agent vessels in their own environments. For every epoch $K$, and for every agent in each environment, try a policy over a set of $\tau$ for an episode of length $T$. Based on the rewards and advantage estimates for given actions, update the policy via (27) by trying to fit the Critic's $V_{\phi_K}$ value against $\hat{R}_t$ using MSE.

**Algorithm 1:** Proximal Policy Optimization (Clipped)

Start with policy $\theta_0$ and value function $\phi_0$ parameters
for $Agent = 0,1, \dots, N_{Agents}$ do
    for $K = 0,1,2 \dots$ do
        ➤ Collect a set of trajectories $D_k = \{\tau_i\}$ by running Actor policy $\pi_K$ over an episode with length $T$.
        ➤ Compute 'rewards-to-go' $\hat{R}_t$ and thus advantage estimates $\hat{A}_t$ based on current value function $V_{\phi_k}$.
        ➤ Update Actor's policy parameters by maximizing the objective function in equation (27) via stochastic gradient *ascent* with Adaptive Moment Estimation (Adam).
        ➤ Fit value function $V_{\phi_k}$ against $\hat{R}_t$ by regression on mean-squared error (MSE) using gradient descent.
    end for
end for

## IV. Object Detection

For the real-time detection of vessels in our virtual environment, we have chosen to use the YOLOX object detection algorithm [19]. Concretely, we use the YOLOX-S (small) model, due to its satisfactory performance and relatively smaller computational cost compared to the larger models, which we convert into the ONNX format [39] for compatibility with the virtual environment. In figure 2 below, we show YOLOX-S's model architecture:

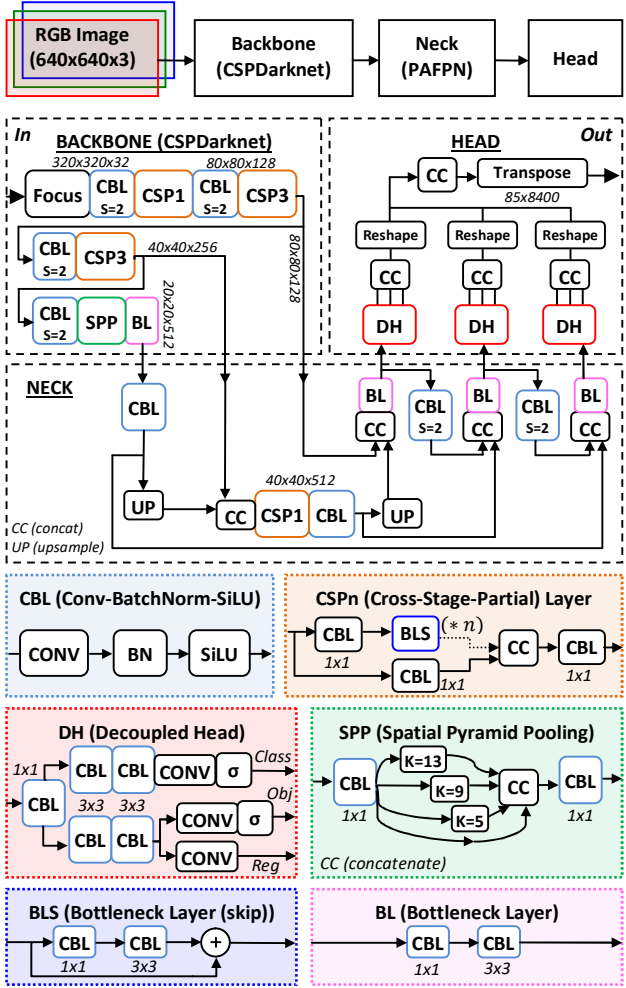

**Figure 2: YOLOX-S model architecture. Adapted from [40].**

In figure 2, all images are resized to the 640x640 pixels required at the input, applying zero-padding in cases where the aspect ratio is not one. The *Focus* layer performs slicing operations on the initial input image tensor followed by concatenation and a CBL layer for down-sampling. The neck module consists of a Path Aggregation Feature Pyramid Network (PAFPN), based on the work of [41]. The activation function used throughout is SiLU (Sigmoid Linear Unit).

YOLOX's main improvements over YOLOv3 [42] are:

- Using a decoupled detector head structure, with three subheads (class (out of 80), objectness, bounding box parameters), for better convergence and performance.
- Incorporating a CSP (Cross-Stage-Partial) backbone based on [43], reducing model complexity.
- Implementing an Anchor-free method, reducing the need for extra hyper-parameters.
- Developing a simplified version of OTA (Optimal Transport Assignment) [44], named SimOTA for label assignment.
- Applying additional data augmentation strategies such as MixUp[45] and Mosaic Augmentation[46].

A combined loss function is calculated after each Decoupled Head, relating to the Class, 'Objectness' and the Bounding Box prediction for each detection before concatenation. For the first two losses Binary Cross Entropy (BCE) loss is used while for the third, Generalized Intersection over Union (GIoU) [47] is used.

## 3. Methodology

All work was conducted using an NVIDIA RTX 2080Ti GPU with 11GB VRAM. The Unity Game Engine (version 2022.2.5f1) was chosen as the environmental simulation platform upon which intelligent maritime Agents were trained using the Unity ML-Agents Toolkit [48], which contains the PPO algorithm. Additionally, we used the Unity Perception Package [49] to generate large-scale synthetic datasets with ground-truth labels that were used to fine-tune the YOLOX-S model, via a curriculum learning strategy inspired from [20], to provide a realistic way for an Agent to sense its surroundings. The YOLOX-S model, in ONNX format, is executed inside Unity using the Unity Barracuda Inference Engine [50] and processed using Compute Shaders to generate bounding boxes for detections on rendered frames.

Therefore, the following sections describe how we set up our simulation environment considering both the visual aspect (ocean rendering, lighting, weather) as well as how we trained our Agents and fine-tuned the YOLOX-S model. We then discuss how the terrain was generated, the developed physics model for each vessel's floating dynamics, how observations were defined for each Agent as well the design of our novel reward functions. Finally, we end this section by describing how we evaluate the performance of the vessels in our environments.

## A. Environmental Setup

### i. Visuals

We use Unity's High-Definition Render Pipeline (HDRP) to render the simulation environment due to its ability to create photorealistic environments through the usage of physically-based lighting techniques. The main visual components of our simulation environment consist of an ocean surface, a physically-based sky with volumetric clouds that can affect ambient and directional lighting levels and weather effects. The digital construction of each is now explained in the context of HDRP.

The ocean model is comprised of a level-of-detail system of multiple concentric square non-overlapping meshes which move in a synchronized manner with the first-person camera view-port. Meshes, each with $256^2$ points, are more densely spaced near the camera than those further away. The horizon distance, $d_H$ by which these furthermost meshes extend is approximated [51] with (31), where $h$ is the height from sea-level:

$$d_H \approx 3570\sqrt{h} \qquad (31)$$

This allows for the illusion of an infinite ocean from the viewers perspective and depending on the value of $h$, meshes can be appended or removed as necessary.

To generate the surface wave mesh, vertices in each of these meshes need be displaced according to the oceanographic theory in section 2.I. of this paper. The calculation of the wave spectra is achieved via the use of DirectX High-Level Shader Language (HLSL) Compute shaders on the GPU. We extract these results onto HLSL RWTexture2D objects which are then passed onto a RenderTexture (RT) to be read by vertex shaders in Unity to displace mesh vertices to specified wave height amplitudes. RT's are mapped onto meshes and repeated across their extent and we use multiple non-periodic wave number ranges in the calculation of the wave spectra, in a manner similar to [23], to avoid tiling artifacts.

In the rendering of the ocean, we incorporate refraction, reflection, subsurface-scattering effects for ambient and directional sunlight while considering Fresnel effects, ocean whitecap formation and rain / snow ripple effects on the waves, as well as weather conditions such as fog, rain and snow.

The first three effects are all part of HDRP's in-built features with the first two being calculated using screen-space reflection and refraction methods. Ripple effects and ocean foam, which is rendered at wave crests using the approach of [21], are added with a custom shader across all ocean waves. The last three weather effects are created and then calibrated for realism using Unity's VFX (Visual Effects Graph) Package.

The sky lighting model is based on [52] and considers the atmospheric scattering of light at all possible viewing angles.

Figure 3 shows a visual ocean scene (TMA wave spectrum) at three different times of day (dawn, midday, evening) in increasingly more challenging weather conditions.

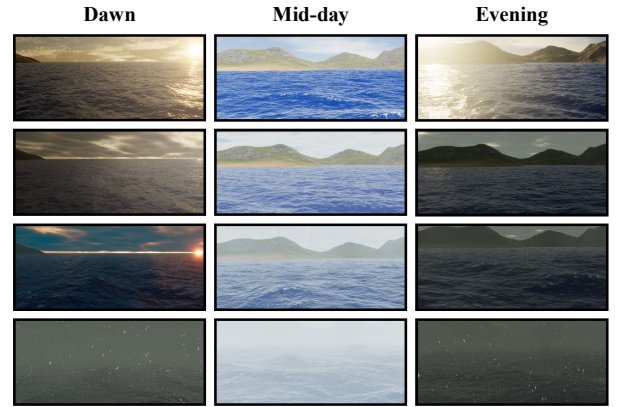

**Figure 3:** Rendering of an ocean scene (TMA spectrum, $U_{10} = 10 ms^{-1}, F = 5km$) at different times of day (left-to-right columns). Images at dawn face in opposite direction of the other two. Weather conditions shown are clear skies (first row), overcast (second row), overcast with rain and light fog (third row) and overcast with snow and heavy fog (last row). Zoom in for more detail.

### ii. Training Procedures

**Training Environments**

In order for our DRL Agents to develop a generalized policy that leads to safe navigational behaviors in unseen situations, we need to able to encapsulate a diverse set of scenarios during the training process within our training environment. The concept of a training environment in our work is shown in figure 4.

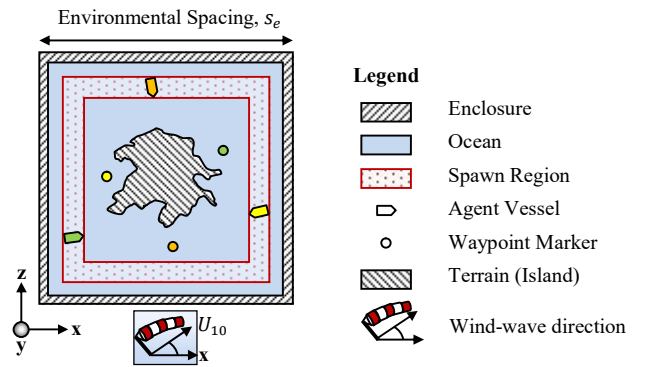

**Figure 4:** Graphic of a training environment and enclosure with associated simulation elements.

In figure 4, multiple agent vessels can exist in a single square training environment of length $s_e$ (typically 500m) with a specified wind-wave direction where each have to navigate towards their own corresponding waypoint marker. At the start of the entire training process, agent vessels are randomly spawned within a user-controlled region. Enclosures are used during training to help speed up convergence. An episode is ended if any of the following conditions are met:

- The agent's proximity to the waypoint is less than 15m.
- The agent has spent more than $T_{Col} = 1500$ time-steps (~half-minute) in collision with another agent or the terrain.
- The agent's episodic time-steps exceed $T_{max}$.

Where:
$$T_{max} = 10 \frac{s_e}{\Delta t \cdot v_{max}} \qquad (32)$$

The term $v_{max}$ is the maximum attainable speed the vessel is capable of (user-input) and $\Delta t$ is the time-step size.

At the end of an episode, the agents resume their navigation towards a newly spawned random waypoint, by starting from the position of the previous waypoint.

The terrain is procedurally generated using random seeding of Perlin Noise [53]. This is explained in more detail in section 3B.

To ensure that we maximize scenario diversity, as well as speeding up training considerably, we extend the concept of a single training environment in figure 4 into multiple training environments. Therefore for the training procedure of our Agents, we form a $4x4$ grid with 2 agents per environment for a total of $N_{vessels} = 32$ Agents being trained concurrently, as Unity ML-Agents Toolkit supports multiple instances of the same agent being trained in parallel.

**Curriculum Strategy for the Object Detection Model**

Apart from training our agents, we also need to fine-tune the YOLOX-S model so that it can provide accurate nearby vessel detections from the EGO vessel's point-of-view. To achieve this, we use python scripts to extract all images in both the Pascal VOC 12 [54] and COCO [55] datasets that contain the class tag "boat" (549 images for the former and 3146 images for the latter). We then compliment these images with our own synthetic images that we generate using our simulation environment with Unity's Perception Package and publically available 3D vessel models [17][56]. In each of the images, all annotations are stored in COCO format. The combined real / synthetic images dataset contains a total of 6226 images of varying resolution, and are randomly shuffled and split with a ratio of $0.8 : 0.1 : 0.1$ for the train, validation and test sets of the dataset respectively.

To generate our synthetic data, we follow a deterministic curriculum strategy similar in some respects to [20] which is summarized in figure 5 below.

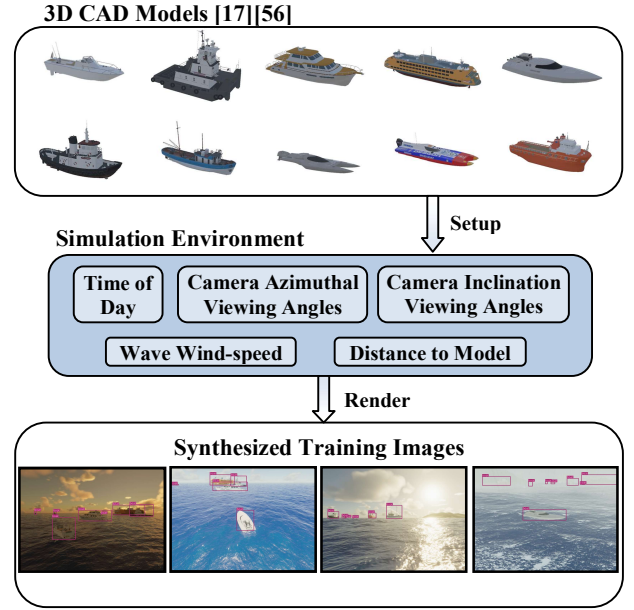

Figure 5: Pipeline for the generation of synthetic training images. The top section shows the 3D Models (not to scale) that are spawned in random poses, each floating on the ocean surface. Within the simulation environment, each of the listed parameters are varied producing rendered scenes as shown in the bottom section with labeled annotations. Zoom in for more detail.

In figure 5, the simulation environment serves as the background for the foreground objects of interest, namely the vessel 3D CAD models. We follow a hierarchical scheduling procedure that includes, in order of execution, the following variations:

1. Azimuthal viewing angles (45° intervals)
2. Inclination viewing angles (at 5° and 25°)
3. Distances to Model (5,20,50 times axial length)
4. Sea-states (Beaufort Scales of 1, 3, 6.5, 9)
5. Times of day (Dawn, Midday, Evening)

Each variation step is repeated by the number of combinations of the stage below it. Therefore, for three different times of day and four sea-states, the camera will render the models in different viewing angles. It should be noted that in all cases, a purely synthetic background (simulation environment) is used for such images.

Figure 6 below also shows selected example training images with bounding box annotations from both the Pascal VOC 12 and COCO datasets.

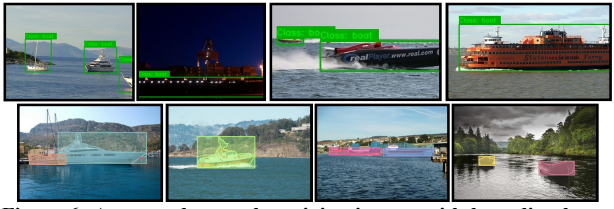

Figure 6: Annotated example training images with bounding boxes from the Pascal VOC (top-row) and COCO (bottom-row) datasets.

We use the YOLOX-S model weights which have been pre-trained on COCO's *train2017* as the starting point for our fine-tuning. We use similar training and data augmentation settings as in the original YOLOX paper [19] and train for an additional 50 epochs. We use a batch size of 8 since we use a single GPU device in this work.

## B. Procedural Terrain

We use Perlin noise [53] to generate our terrain. Specifically, we introduce a falloff functions onto this noise to transform the terrain into notional islands with coastlines and underwater reefs with a grid height of $\eta_{PN}$. The perlin noise itself, *Mathf.PerlinNoise* ( *PN* ), is calculated using the C# .Net mathematics library. The process for this is shown in pseudo-code format in Algorithm 2:

---
**Algorithm 2:** Perlin Noise for Procedural Islands
---

**for** $i = 0,1, \ldots, N$ **do**
    **for** $j = 0,1, \ldots, N$ **do**
        $f_v = \max(\frac{2i}{N} - 1, \frac{2j}{N} - 1)$
        $f_s = \frac{f_v^{(f_A)}}{f_v^{(f_A)} + (f_B - f_B f_v)^{(f_A)}}$
        $\eta_{PN} = 0, \eta_A = 1, \eta_\omega = 1, \eta_p = 0.5, \eta_l = 2.5$
        **for** $o = 0,1, \ldots, N_{octaves}$ **do**
            $(i_s, j_s) = \frac{(i,j) - \frac{N}{2}}{\eta_{scale}} * \eta_\omega * r_{offset}$
            $noise = 2PN(i_s, j_s) - 1$
            $\eta_{PN} \mathrel{+}= noise * \eta_A$
            $\eta_A \mathrel{*}= \eta_p, \eta_\omega \mathrel{*}= \eta_l$
        $\eta_{PN} \mathrel{-}= f_s * f_m$
        $\eta_{PN,Final} = \eta_{PN} * \eta_m$

---

Where $f_v$, the fall-off value is used in $f_s$ to provide a smooth variation in height drop-off the further away from the terrain centre. $N_{octaves}$ controls the levels of perlin noise to be applied. Terms $f_m$ and $\eta_m$ are the falloff and height multipliers respectively, while $\eta_{scale}$ is a parameter that controls the smoothness of the terrain surface. Term $r_{offset}$ can be used as an additional randomization parameter. Larger values of $f_A$ and $f_B$ result in a more rectangular shape for the terrain, while smaller values lead to a pyramidal shape. Lastly, $\eta_p$ and $\eta_l$ are the persistence and lacunarity parameters. The former controls the consecutive influence of the level of perlin noises for each octave with $\eta_p < 1$ leading to reduced influence. The latter, if increased, leads to more surface roughness.

We use values of $f_A = f_B = 2$, $\eta_{scale} = 50$, $N_{octaves} = 4$, $\eta_p = 0.5$, $\eta_l = 2.5$, $f_m = 5$, $\eta_m = 15$, and $r_{offset} = 1337$.

## C. Vessel Dynamics and Propulsion Models

A critical aspect of the simulation is the dynamical interaction that vessels have with the simulation environment.

Unity's Physics Engine integrates the Nvidia PhysX Engine [57], which allows for the efficient simulation of rigid-body dynamics as well as collisions. Thus the focus of this part of the work was on developing the necessary mathematical dynamical models for marine vessels that would allow for realistic vessel-water interactions within such a Physics Engine.

All 3D vessels in figure 5 and the ocean are geometries discretized into subdivided triangular facets that are connected to each other, each comprised of edges and vertices. Both vessel and ocean vertices exist within Unity's Left-Handed 3D coordinate system. The starting point for calculating vessel-water interactions is extracting the difference of the vertical positions (Y axis) of these two sets of vertices at approximately the same horizontal location (X, Z axes).

As described in section 3A.i., the ocean surface height is stored in a RenderTexture. We load the height values from the RT into a Texture2D object on the GPU using Unity's "*Texture2D.LoadRawTextureData*" command, and then perform an optimized bilinear filtering operation via "*Texture2D.GetPixelBilinear*". This allows us to extract the vertical position of an ocean vertex at approximately the same horizontal position as the vessel vertex. We thus calculate the vertical difference between these two positions, with a positive difference meaning above water and negative meaning submerged. In this way we can decide on the appropriate environmental forces to apply on each triangle.

The types of environmental forces we incorporate into our physics model and apply on the vessels are:
- Hydrostatic (Buoyancy) Forces
- Hydrodynamic Drag and Friction Forces
- Aerodynamic Drag Forces
- Local Wind-driven Water Current Forces

To describe each of these forces in detail, figure 7 provides the necessary graphical context and notation:

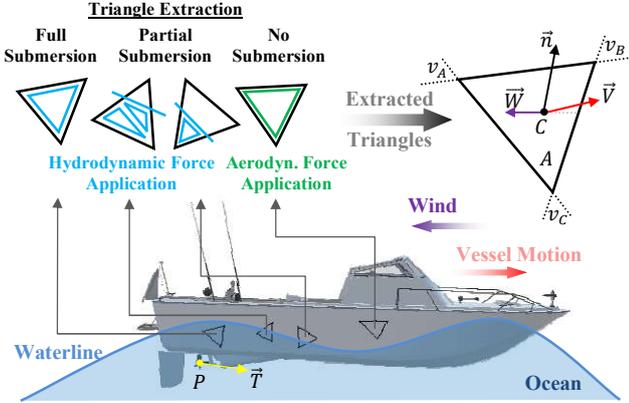

Figure 7: A schematic showing the EGO vessel model [17] (bottom) immersed in water at different levels across the hull surface. Each vessel mesh triangle (black) is comprised of vertices $v_A$, $v_B$ and $v_C$ in the world-frame coordinate system. If the waterline does not intersect the triangle, hydrodynamic forces (in blue) or aerodynamic forces (in green) are applied across the full triangle. Otherwise, the submerged part of an intersected triangle is sub-divided and hydrodynamic forces are applied on those extracted triangles.

For every extracted triangle illustrated in figure 7, the term $A$ is the area of triangle, $C$ the geometric center at a vertical (always negative) distance $C_h$ from the water surface calculated using barycentric coordinates, $\vec{V}$ is the velocity vector of the triangle, $\vec{n}$ is the outward pointing normal and $\vec{W}$ the wind direction in unit vector form, which can point either inwards or outwards of the triangle surface.

We begin by defining $\vec{V}$ via equation (33):

$$\vec{V} = \vec{U} + (\vec{\Omega} \times \overline{CG}) \qquad (33)$$

Where $\vec{U}$ and $\vec{\Omega}$ are the linear and angular velocities of the vessel respectively, while $\overline{CG}$ is the vector from the vessel center of mass coordinate, $G$, of the vessel to the triangle center.

From equation (33), we can start calculating our modeled forces. For every submerged triangle, the hydrostatic force $\vec{F}_B$ is calculated by:

$$\vec{F}_B = \rho_{sea} A C_h g (\vec{n} \cdot \hat{y}) \hat{y} \qquad (34)$$

Where $\hat{y}$ is a unit vector pointing upwards in the direction of the y-axis. The dot product between $\vec{n}$ and $\hat{y}$ is used to account for the vertical area component of the triangle.

The hydrodynamic drag force, $\vec{F}_D$, where $\overrightarrow{V_n} = \frac{\vec{V}}{|\vec{V}|}$, is:

$$\vec{F}_D = -\rho_{sea} A (\vec{n} \cdot \overrightarrow{V_n}) n \left( |\vec{V}|^q \right) C_D(Re) \qquad (35)$$

In equation (35), $q \geq 1$ and it defines whether drag scales linearly ($q = 1$) or non-linearly ($q > 1$). We use $q = 1$. The term $C_D(Re)$ is an *approximate* Coefficient of Drag variable that is a function of the Reynolds Number, $Re$ given by:

$$Re = \frac{\rho_{sea} |U| L}{\mu_{sea(20C°)}} \qquad (36)$$

Here, $\mu_{sea(20\ °)} = 0.00109\ Nsm^{-2}$, is the dynamic viscosity of sea-water at $20C°$, $L$ is the axial length of the vessel and $|U|$ is the velocity magnitude of the vessel. With this value of $Re$ we can evaluate $C_D(Re)$ based on the graph shown in figure 8 below from experimental data according to [58]:

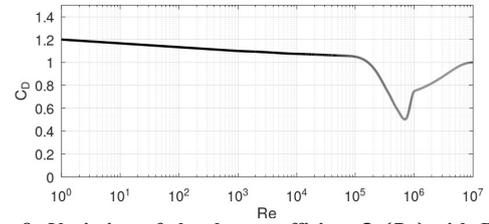

Figure 8: Variation of the drag coefficient $C_D(Re)$ with Reynolds Number, $Re$.

In (35) if the dot product of $n$ and normalized $\vec{V}$ is positive then the drag force pushes the vessel, otherwise, if negative, it pulls it in the given direction of $\vec{V}$.

The hydrodynamic friction force and local wind-driven current forces are combined into $\vec{F}_{FC}$, which is:

$$\vec{F}_{FC} = \rho_{sea} A s_F (\vec{W} w_C U_{10} - \vec{V}) \qquad (37)$$

Where:

$$s_F = (1 - \vec{n} \cdot \overrightarrow{V_n})(\vec{n} \cdot \overrightarrow{V_n}) C_F(Re)$$

The term $w_C$ is a user controlled current factor that controls the viscous drag from the waves. The term $C_F(Re)$ is the *approximate* Coefficient of Friction, defined by (38) from [59]:

$$C_F(Re) = \frac{0.075}{(log_{10}(Re) - 2)^2} \qquad (38)$$

Each of the aforementioned forces apply for submerged triangles. Summed together, they lead to a total force, $\vec{F}_T$ which is then applied at point $C$ of each triangle:

$$\vec{F}_T = \vec{F}_B + \vec{F}_D + \vec{F}_{FC} \qquad (39)$$

Lastly, for all non-submerged triangles the aerodynamic drag force, $\vec{F}_{AD}$, also applied at point $C$, is calculated by:

$$\vec{F}_{AD} = -\frac{1}{2} \rho_{air} A U_{10}^2 \vec{W} (\vec{n} \cdot \vec{W}) \qquad (40)$$

The density of air $\rho_{air}$ at sea-level is $1.225 kgm^{-3}$.

Therefore, depending on whether a mesh triangle is submerged or not, the forces $\vec{F}_T$ or $\vec{F}_{AD}$ that move the rigid-body are also applied on a corresponding position vector $\overline{CG}$ to generate torque. The 6-DOF motion that this rigid-body undergoes from the application of these forces is handled by Unity's physics engine.

The physics model described through equations (33-40) is scale agnostic – it can be applied in the same way to jet skis all the way up to cargo ships - as long as care is taken in choosing of appropriate mesh resolutions. In our physics model, all calculations are run using multi-threading code via Unity's Job System (Burst Compiler).

The propulsion system is based on applying a propulsive thrust force $\vec{T}$ at point P in figure 7, generating both linear and angular accelerations from moments that induce motion. The direction of $\vec{T}$ is in the vessels frame-of-reference with its zero-yaw, zero-trim position facing the forward axial direction of the vessel. Multiple propulsion systems may also be added onto a vessel.

As this propulsion system is the direct way in which a vessel agent will interact with its environment, its continuous action space vector $A$, as described in section 2.III., will consist of two components (both normalized between $-1$ and $1$); a thrust magnitude $|\vec{T}|$ and a yaw angle input, $T_y$ relative to the forward axial direction:

$$A = [|\vec{T}|, T_y]^T \qquad (41)$$

The thrust magnitude $|\vec{T}|$ is approximated by:

$$|\vec{T}| \approx \frac{\eta_p Q}{|\vec{U}|}$$

Where we assume $\eta_p \approx 0.75$ to be the total propulsion efficiency and $Q$ the propulsive engine power in Watts.

### D. Agent Observations

The Agents are aware of the environment state-space via an observation space vector, $O$ which is comprised of the following elements; a binary collision observation feature $o_{Col}$, a waypoint observation feature vector $o_{WP}$, as well as static and dynamic obstacle perception feature vectors, $o_S$ and $o_D$ respectively:

$$O = [o_{Col}, o_{WP}, o_S, o_D]^T \qquad (42)$$

Each observation feature in expression (42) is obtained at every time-step, $t$.

We begin with the simplest feature, $o_{Col} \in \{0,1\}$ which is 1 when the vessel is in contact either with terrain or another vessel, 0 otherwise. The aspect of collision itself is handled through Unity's physics engine. This allows the Agent to 'learn' through previous experiences that all actions that lead, during training, to a value of $o_{Col} = 1$ should be avoided.

The waypoint observation feature vector, $o_{WP}$ is comprised of two parts. The first is a normalized heading error, $\beta \in [-1,1]$, which is calculated by obtaining the signed angle (in degrees) between the vessel's normalized axial-forward direction vector, $\vec{x}_{fwd}$ of the vessel and the normalized direction vector pointing to the current waypoint, $\vec{x}_{wp}$. The second part is a normalized distance to the current waypoint, clamped between [0,1], $d_{WP^*} \in [0,1]$. It is the ratio of the distance of the vessel to the waypoint at the current time-step, $d_{WP}^t$ over the distance of the vessel to the waypoint at the start of the current episode, $d_{WP}^{\tau(t=0)}$. Thus:

$$o_{WP} = [\beta, d_{WP^*}]^T \qquad (43)$$

Where:

$$\beta = \frac{cos^{-1}\left(\frac{\vec{x}_{fwd} \cdot \vec{x}_{wp}}{|\vec{x}_{fwd}||\vec{x}_{wp}|}\right)}{180} \ , \ d_{WP^*} = d_{WP}^t / d_{WP}^{\tau(t=0)}$$

While other approaches in the literature [15] have adopted observation features that monitor how well an Agent follows a pre-determined path, the existence of significant wave and wind disturbances present in this work may sometimes lead to negative rewards being applied onto an Agent from a temporary deviation of the path, by no fault of its own. This is why in (43), we adopt fewer observation features in regards to the navigation towards waypoints as they are more easily related (and affected) by the possible acceleration and steering actions chosen by the Agent at any given time-step.

Both feature vectors $o_S$ and $o_D$, are based on phenol-menologically modeling rangefinder sensors (such as active sonar for $o_S$ and radar for $o_D$) via Unity's "*Physics.Raycast*" function that uses ray-triangle intersection calculations for terrain and vessels. The interested reader can see [60] for more information on this calculation procedure.

In the case of static obstacle detection we model, as a first-order approximation, the transmission-reception of the Sonar echo by casting mathematical rays from the center of mass of the sensing vessel outwards across 360° degrees in equally spaced intervals of 7.2° for a total of $N_{rays} = 50$ rays over a total effective distance of $D_{Sonar} = 150m$. All rays are casted at a slightly negative angle ($-4°$) and return the following information, $\zeta$ as shown in (44):

$$\zeta = \{\vec{\zeta}_{ray}, \zeta_{hit}, \zeta_{frac}\} \qquad (44)$$

Where $\vec{\zeta}_{ray}$ is normalized ray vector, $\zeta_{hit}$ is a Boolean

which is 1 if the ray collided with terrain, 0 otherwise, and $\zeta_{frac}$ is the fraction of $D_{sonar}$ at which the collision took place otherwise 1. With this information, it is then trivial to calculate the angles of these rays, $\theta_{ray}$ relative to vector $\vec{x}_{fwd}$.

In the case of dynamic obstacle detection we also model, as a first-order approximation, the transmission-reception of radar waves via the casting of a mathematical ray from the center of mass of the sensing vessel that rotates around the vertical y-axis with $\omega_{radar}$ at 60 rotations per minute (rpm). The total effective distance in this case is $D_{Radar} = 500m$. In cases when the ray collides with another target vessel, the collision distances, $d_{\vec{v}_\#}$ and heading angles, $\theta_{\vec{v}_\#}$ (in degrees) relative to the EGO vessel's position and $\vec{x}_{fwd}$ vector are normalized using $d_{WP}^{\tau(t=0)}$ and $1/180$ respectively and stored in cache. Furthermore, the velocities of each detected target vessel, $\vec{v}_\#$ are directly extracted from the simulation environment and decomposed into incoming and orbital speeds ($v_{i,\#}$ and $v_{o,\#}$) relative to the sensing vessels position in a similar manner to how it was achieved in [14]. These are normalized by dividing by the vessels maximum attainable propulsion speed, $U_{max}$ set by the user. Target vessel detection data is thus summarized by (45):

$$\psi = \{d_{\vec{v}}, \theta_{\vec{v}}, v_i, v_o\} \quad (45)$$

Figure 9 below shows how these ray-casts were used for both static and dynamic obstacle perception:

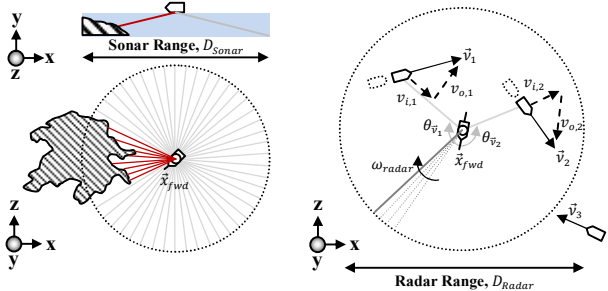

**Figure 9:** Phenomenological sonar (left) and radar (right) models using mathematical ray-casting. In the case of static obstacle detection (left), red lines indicate collisions with the terrain at different fractions of total range, $D_{Sonar}$. For dynamical obstacles, two target vessels with velocities $\vec{v}_1, \vec{v}_2$ within radar range $D_{Radar}$, have their velocities decomposed into incoming (facing sensing vessel) and orbital (tangent to sensing vessel) velocity components. Target vessel with $\vec{v}_3$ is not considered as it has not entered $D_{Radar}$. Sign convention for these velocities is positive for motion toward the sensing vessel, and clockwise orbital motion. Two separate scenarios. Not to scale.

The static and dynamic obstacle perception feature vectors, $o_S$ and $o_D$ can therefore be expressed by (46) and (46). For $o_S$, for each ray indexed from 1 to $N_{rays}$, we have:

$$o_S = [\zeta_1, \zeta_2, \ldots, \zeta_{N_{rays}}]^T \quad (46)$$

For $o_D$, for each detected vessel starting at index 1, up to a total of $N_{vessels}$, we have:

$$o_D = [\psi_1, \psi_2, \ldots, \psi_{N_{vessels}}]^T \quad (47)$$

The first two terms of $\psi$ in (45) (i.e. target distance and heading) depend on significantly on the value of $\omega_{radar}$. If this value is too low, the location and velocity of a target vessel detected at the start of one radar sweep may be considerably different by the time of the next sweep. Values larger than 60 rpm can lead to target vessels being missed entirely since the rotation of the ray-cast takes place over a fixed time-step size. Therefore, to robustly improve target distance and heading readings, we perform sensor fusion by combining the detections from the radar model with the object detection model that was discussed and fine-tuned in section 3.A.ii.

This is done with the "*Camera.ScreenPointToRay*" function in Unity, which takes as input a pixel coordinate on the screen and performs a ray-cast in the direction of that pixel from the EGO vessel's position and point-of-view. In our case, such pixel coordinates are chosen to be the center of bounding box detections generated using the YOLOX-S model. Upon collision, this "object detection ray-cast" returns real-time the position of the virtual object it collided with in the simulator.

Finally, it should be noted that every vessel present in the simulation has its own concatenated observations feature vector $O$ as described in (42), with the exception of the object detection sensor-fusing stage which applies only to the EGO vessel. To generate all these observations efficiently as the number of vessels scale upwards we again use multi-threading code via Unity's Job System (Burst Compiler).

**E. Agent Rewards**

Agents are rewarded (or penalized) based on the actions they decided to take within their environment from the observations they receive. By designing appropriate reward functions we can guide the training of an Agent to follow the required collision avoidant navigational behavior.

In our work, we extract a scalar reward signal continuously at each time-step, $R_t$ as well as once at the end of an episode, $R_{t=\tau}$. Both rewards are defined in expressions (48) and (49):

$$R_t = \begin{cases} r_{Total} & , o_{Col} = 0 \\ r_{Col} & , o_{Col} = 1 \end{cases} \quad (48)$$

$$R_{t=\tau} = \begin{cases} r_{WP} & \begin{cases} o_{WP} \to [0,0]^T \\ o_{Col} = 0 \end{cases} \\ r_{Col,\tau} \quad , & o_{Col} = 1 \end{cases} \quad (49)$$

Negative collision rewards $r_{Col} = -0.25$ and $r_{Col,\tau} = -3$ are applied during collision, where $o_{Col} = 1$. In a scenario where the agent has collided with terrain or another vessel and no remedial action is taken then, as described in section 3.A.ii, the negative cumulative reward for that episode would be $T_{Col} * r_{Col} = -375$.

Upon reaching a specified waypoint, the Agent is rewarded with a fixed value reward of $r_{WP} = 3$ and the next episode begins.

The largest contribution to the reward signal however comes from the combined reward $r_{Total}$ applied every time-step, as shown in (50):

$$r_{Total} = r_N + r_S + r_D + r_X \quad (50)$$

Here, $r_N$, $r_S$ and $r_D$ are the navigational, static and dynamic collision avoidance reward components respectively obtained at every time-step. The term $r_X$ is a time penalty defined by $r_X = -T_{max}^{-1}$ that encourages the Agent to achieve its goals within a reasonable time-frame. The first three terms are now described.

**Navigational Reward, $r_N$**

In order to encourage the Agent to navigate towards a specified waypoint, reward component $r_N$ is constructed by incorporating the linear velocity vector of the vessel, $\vec{U}$, the vessel's normalized axial-forward direction vector, $\vec{x}_{fwd}$, and the normalized direction vector to the waypoint, $\vec{x}_{wp}$. Given this information, two dot-products, $\hat{U} = (\vec{U} \cdot \vec{x}_{wp})$ and $\hat{H} = (\vec{x}_{fwd} \cdot \vec{x}_{wp})$ can then be used to gauge how well the vessel is progressing towards its waypoint as well as its alignment to it respectively. For clarity on the usage of these dot-products, figure 10 outlines the four possible cases of a vessel's progress towards a waypoint:

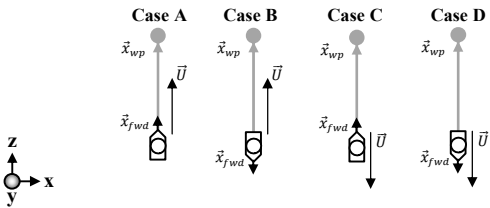

**Figure 10:** Vessel progress towards a waypoint (grey circle). In cases A and C there is vessel alignment ($\hat{H} > 0$) but opposing resulting vessel velocities, whereas in cases C and D, the vessel is moving away from the waypoint ($\hat{U} < 0$). To maximize $r_N$, both dot-products must be positive and as close to one as possible. In case D, the product $\hat{U}\hat{H}$ is made negative through the use of an *if* statement.

Reward component $r_N$ is given by:

$$r_N = r_r U_f \Omega \hat{U} \hat{H} \quad (51)$$

In equation (51), the term $r_r = 0.1$ is a user-controlled scaling parameter for the reward signal and was chosen through trial-and-error. The vessel velocity factor, $U_f$ is used to discourage vessel speeds near the maximum attainable propulsion speed, $U_{max}$ via user-set a velocity margin parameter $U_{up} \approx U_{max}/5$. $U_f$ is:

$$U_f = \frac{0.2}{1+e^{(0-(U_{max}-U_{up}))}} \quad (52)$$

The term $\Omega \in [0,1]$ is a weighting factor that balances between reward components $r_N$ and $r_D$, where in the latter case it is multiplied as an inverse term, $(1 - \Omega)$. In cases where incoming vessels approach the EGO vessel, $\Omega$ will tend towards zero. In this way, by reducing the impact of $r_N$ and consequently increasing $r_D$, the dynamic collision avoidance reward signal will take precedence over progressing towards a waypoint. This concept of a trade-off between reward signals has previously been applied in [15], however in our work the calculation of this weighting factor incorporates the concept of collision time, $t_c$ and $t_{c,s}$. The term $t_c = d_{\vec{v}}/(|\vec{U}|_{v_i} + v_i)$ is the time (in seconds) an imminent collision will take place with each detected vessel $\psi$ (from (45)) at the current time-step, where $|\vec{U}|_{v_i}$ is the speed towards the detected vessel. The term, $t_{c,s} = 20$, is a user-step value (in seconds) which signifies how many *safety* seconds in advance the trade-off between reward signals should begin. $\Omega$ is therefore defined by the minimum value of $t_c$ for all detected vessels defined by $\psi$ in observation feature vector $o_D$, as shown in (53):

$$\Omega = \min_{t_c \forall (\psi \in o_D)} \left[ \frac{1}{1 + e^{\left(-\frac{1}{2}(t_c - t_{c,s})\right)}} \right]$$

(53)

**Static Collision Avoidance Reward, $r_S$**

The Agent must avoid running aground during navigation towards waypoints. To encourage the Agent in heeding this critical behavior, we implement a static collision avoidance reward, $r_S$ which at best should tend towards zero, otherwise have a negative value and therefore performing as a penalty signal. To construct $r_S$ we need to use the terrain information observation feature vector $o_S$ from (46). For each ray that has collided with terrain, as shown in figure 9, we extract their angle $\theta_{ray}$ relative to $\vec{x}_{fwd}$, and obtain the ray length, $\zeta_L$ from $\zeta_L = \zeta_{frac} D_{Sonar}$. For a given vessel speed $|\vec{U}|$, we can therefore estimate the time (in second), $t_{Land}$, the current vessel will run aground for each ray via equation (54). Note that the denominator, also denoted as $U_{Land}$, is clamped between a small and large positive value:

$$t_{Land} = \frac{\zeta_L}{[|\vec{U}||\cos(\theta_{ray})|]_{10^{-3}}^{10^3}} \quad (54)$$

Therefore, with a penalty scaling term $p_r = 0.05$, equation (55) defines the static collision avoidance reward, $r_{S,ray}$ for each ray:

$$r_{S,ray} = p_r U_{Land} e^{\left(-\frac{t_{Land}}{U_{Land}}\right)} \quad (55)$$

Since the reward signal $r_S$ needs to be a scalar however, we need somehow to combine each $r_{S,ray}$ term. We also need to consider the alignment of the vessel given by each collided ray angle $\theta_{ray}$ (in degrees), since terrain in front or behind the vessel is more dangerous than at its side due to the surging motion generated by the propulsion system. To this end, we specify $H_{Land}$ which is a weighting factor based the heading of the vessel towards land, as shown in (56):

$$H_{Land} = 1 + \frac{2}{1+e^{\left(\frac{1}{2}|\theta_{ray}|\right)}} - \frac{1}{1+e^{(-\theta_{ray}-180)}} + \frac{1}{1+e^{(180-\theta_{ray})}} \quad (56)$$

This is then used to produce a weighted average by summing over every collided ray, yielding the overall $r_S$ value:

$$r_S = -\frac{\sum_{ray=1}^{Nrays}(H_{Land} r_{S,ray})}{\sum_{ray=1}^{Nrays}(H_{Land})} \quad (57)$$

In situations where no ray has collided with any terrain, $r_S$ is set to zero ($r_S = 0$).

**Dynamic Collision Avoidance Reward, $r_D$**

The Agent can learn to avoid other vessels in a COLREG-compliant manner during navigation towards waypoints via the implementation of a dynamic collision avoidance reward, $r_D$. Central to the construction of $r_D$ is the heading $\theta_{\vec{v}}$ of each nearby detected target vessel, from detections $\psi$ in feature vector $o_D$. This is because, as COLREGs dictate, motor vessels should pass each other from the port side to the stern, avoiding the starboard side where possible. Figure 11 shows these different regions:

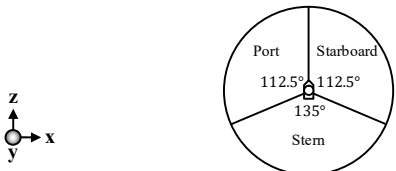

**Figure 11: Terminology for each region around the vessel.**

Concretely, $r_D$ is a scalar and can either be a positive or negative. It is the summation of every dynamic collision avoidance reward per detected target vessel, $r_{D,\psi}$. Regarding each vessel, the term $r_{D,\psi}$ consists of the following parts; a heading COLREG influence factor $H_C$, a distance penalty term, $r_d$ and an orbital reward term $r_o$. Reward component $r_D$ is given by:

$$r_D = \sum_{\psi=1}^{N_{vessels}}(r_{D,\psi}) \quad (58)$$

Each detected vessel, $\psi$ from the EGO vessel's perspective leads to an avoidance reward $r_{D,\psi}$ defined by:

$$r_{D,\psi} = -(H_C r_d) + r_o \quad (59)$$

Since each of the above terms use Sigmoid functions heavily, for clarity we first show it in equation (60). We also use some calibrating parameters, $S_1, S_2, S_3$, to the function for control:

$$\sigma_{(S_1,S_2,S_3)} = \frac{S_1}{1+e^{-S_2(x+S_3)}} \quad (60)$$

Thus, the COLREG influence factor, $H_C$, is:

$$H_C = 1 - \sigma_{(1,\frac{1}{4},-112.5)} + \sigma_{(\frac{7}{4},\frac{1}{4},5)} - \sigma_{(\frac{3}{4},\frac{1}{4},112.5)} \quad (61)$$

The distance penalty term, $r_d$, incorporates $(1 - \Omega)$ as discussed previously, the combined (and clamped) speed between EGO vessel and detected target, $U_\psi$, and a distant COLREG influence factor, $H_{C,r_d}$ that is used to increase the impact of COLREGs based on $d_{\vec{v}}$ from $\psi$. Therefore $r_d$ is:

$$r_d = (1 - \Omega) p_r U_\psi e^{\left(-\frac{0.5 d_{\vec{v}}}{U_\psi H_{C,r_d}}\right)} \quad (62)$$

Where:

$$U_\psi = [|\vec{U}|\cos(\theta_{\vec{v}}) + v_i]_{10^{-3}}^{10^3}$$

$$H_{C,r_d} = 1 - \sigma_{(1,\frac{1}{4},112.5)} + \sigma_{(3,1,5)} - \sigma_{(2,1,-5)}$$

Lastly, the orbital reward term $r_o$ is used to encourage the Agent to adopt navigational patterns where target vessels pass by it with a positive (clamped) anti-clockwise tangential orbiting velocity, $[v_o]_{10^{-3}}^{10^3}$, recalling figure 9 and equation (45). An orbital COLREG influence factor $H_{C,r_o}$ is also used to aid in this. The term $r_o$ is:

$$r_o = r_r v_o H_{C,r_o} e^{\left(-\frac{0.15\vec{v}}{v_o}\right)} - \frac{2v_o}{v_o + e^{(|0.25 d_{\vec{v}}|)}} \quad (63)$$

Where:

$$H_{C,r_o} = \sigma_{(1,\frac{1}{4},112.5)} - \sigma_{(1,\frac{1}{4},5)}$$

In situations where no target vessel in within range of the sensors, $r_D$ is set to zero ($r_D = 0$).

## F. Methodological Parameters

Table 1 below provides a summary of each of important parameters and hyper-parameters used throughout this work. It should be noted that across the PPO network with ML-Agents, the Swish activation function is used. Also, since multiple Agents are used during training, each step taken by these agents cumulatively contribute to the maximum number of steps trained, $K$.

**Ocean &Vessel Physics Models**

| Symbol | Value | Description |
|---|---|---|
| $\rho_{air}$ | $1.225\ kgm^{-3}$ | Density of air at sea-level |
| $\rho_{sea}$ | $1025\ kgm^{-3}$ | Density of seawater ($20C°$) |
| $D$ | $150\ m$ | Sea Depth |
| $L$ | $\sim 8\ m$ | Axial Length of EGO Vessel |
| $\mu_{sea(20C°)}$ | $0.00109\ Nsm^{-2}$ | Dynamic viscosity of sea-water ($20C°$) |
| $U_{max}$ | $25\ ms^{-1}$ | EGO Vessel maximum speed |
| $F$ | $5\ km$ | JONSWAP Fetch |
| $Q$ | $300HP$ | Propulsive Engine Power |

**Proximal Policy Optimization (PPO) with ML-Agents**

| Symbol | Value | Description |
|---|---|---|
| $N_{vessels}$ | 32 | Number of Agents trained in Parallel |
| $\lambda$ | 0.0003 | Learning Rate |
| $\gamma$ | 0.995 | Discount Factor |
| $\epsilon$ | 0.1 | PPO Clipping Value |
| $\lambda_{PPO}$ | 0.99 | Regularization Value for GAE |
| $\beta_{PPO}$ | 0.01 | Policy Randomization Strength |
| $n_{layers}$ | 3 | Number of Hidden Layers in Network |
| $n_{units}$ | 1024 | Number of Hidden Units per Layer in Network |
| $n_{batchsize}$ | 4096 | Number of Agent time-step experiences per iteration of gradient descent |
| $n_{buffersize}$ | 65536 | Number of Agent time-step experiences to collect before updating policy |
| $K$ | $2 \times 10^8$ | Maximum combined number of training steps |
| $T_{max}$ | $10^4$ | Maximum allowable steps per episode |
| $\tau_{PPO}$ | 1024 | Time Horizon Parameter |

**Observation Parameters & Reward Signals**

| Symbol | Value | Description |
|---|---|---|
| $D_{Radar}$ | $500\ m$ | Effective range for radar-like model |
| $\omega_{radar}$ | $60\ rpm$ | Radar Sweep |
| $D_{Sonar}$ | $150\ m$ | Effective range for sonar-like model |
| $N_{rays}$ | 50 | Number of ray-casts for sonar-like model |
| $r_r$ | 0.1 | Reward Scaling Term |
| $p_r$ | 0.05 | Penalty Scaling Term |
| $t_{c,s}$ | $20\ s$ | Safety Collision Time |
| $O_{conf}$ | 0.5 | Object Detection Model Confidence Threshold |

**Table 1: Summary of important parameters used throughout the Simulation Environment.**

## G. Performance Evaluation

Agent performance is evaluated in two stages. The first stage tests for COLREG-compliance in each scenario outlined in Figure 1 for rules 13, 14 and 15 with two vessels, both Agents, and away from any coastline. In the second stage, the behaviors of the Agents are observed in proximity to procedurally generated islands and with additional nearby vessels, to test qualitatively how well COLREG-compliance to the rules is upheld in progressively more challenging situations, such as those of decreased visibility and weather conditions.

The performance and accuracy of the fine-tuned Object Detection model is evaluated separately from the Agent scenarios. In this case, we record videos under different visibility and weather conditions to evaluate how accurately the predicted bounding boxes compare with their ground-truth counterparts. To do this we use the average precision ($AP$) metric shown in (67) which is the area under the precision ($PR$) vs. recall ($RE$) curve:

$$IoU = \frac{TP}{TP+FP+FN} \quad (64)$$

$$PR = \frac{TP}{TP+FP} \quad (65)$$

$$RE = \frac{TP}{TP+F} \quad (66)$$

$$AP = \int_0^1 PR\ (RE)\ d(RE) \quad (67)$$

To deem whether a prediction is considered a true-positive ($TP$), a false-positive ($FP$) or a false-negative ($FN$), we use an $IoU$ threshold of $0.5$. Thus, for detections in a single frame, it will be a $TP$ if $IoU > 0.5$, a $FP$ if $IoU \leq 0.5$ and a $FN$ if no prediction is made at all. We collect such frames every 10 frames of video (~ 1 minute) and average all $AP$ values from each one of these frames for a final evaluation metric, $AP_V$.

## 4. Results and Discussion

We first show the training history for the Agents in figure 12. Each value on the x-axes signify an episode of length $T_{max}$ or less each $N_{vessels}$ experienced in parallel to form a combined number of experiences of steps, $K$. On the y-axes, the averaged cumulative reward signal and average episode length ratio is calculated across the 32 Agents.

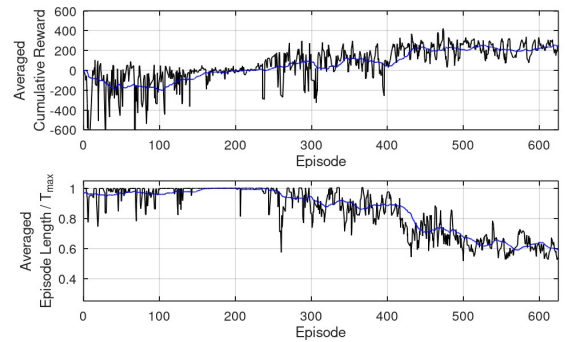

**Figure 12: Training histories (black) with moving averages (blue). Averaged cumulative reward (top) and episode length (bottom) extracted at the end of each Agent's episode.**

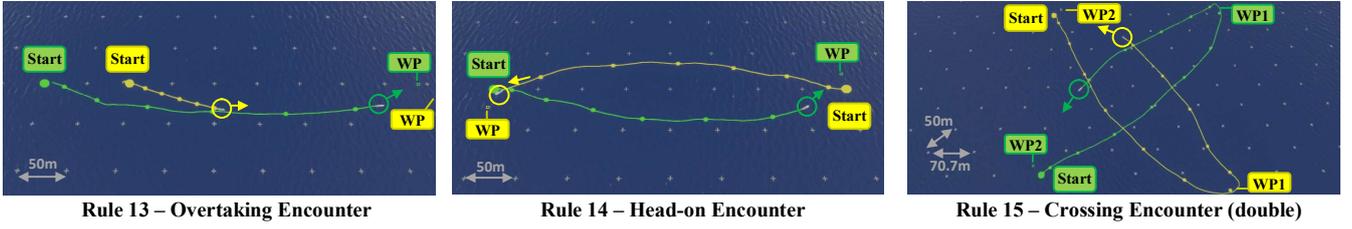

**Figure 13:** COLREG-compliance tests for Rules 13 (left), Rule 14 (center) and Rule 15. For the vessels tested, Rule 16 is also observed. Gray crosses show distance. Green and yellow lines signify trajectories of two separate Agents navigating in awareness of each other in the simulation environment. A single large sphere denotes the starting point and each consecutive smaller sphere measures a time of 10 seconds. 'WP' is Waypoint. Foam effects are removed for clarity. Ocean settings: TMA, $U_{10} = 4ms^{-1}$ (westwards) and $F = 5km$.

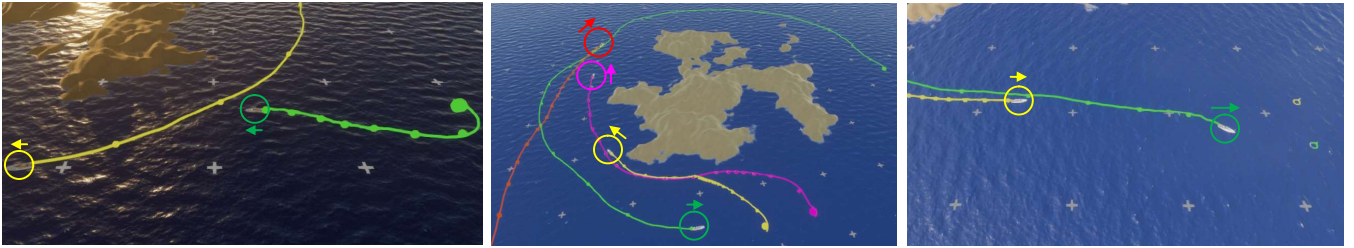

**Figure 14:** Combined COLREG-compliance tests in different randomized training scenarios for Rules 13, 14, 15. The same notation is used as in figure 13 for waypoints, trajectories, colors and time-stamping. In all situations, the Agents safely avoid the terrain. Foam effects are removed for clarity. Ocean settings: TMA, $U_{10} = 6ms^{-1}$ (westwards) and $F = 5km$. Horizontal distance between gray crosses is $50m$, as in figure 13.

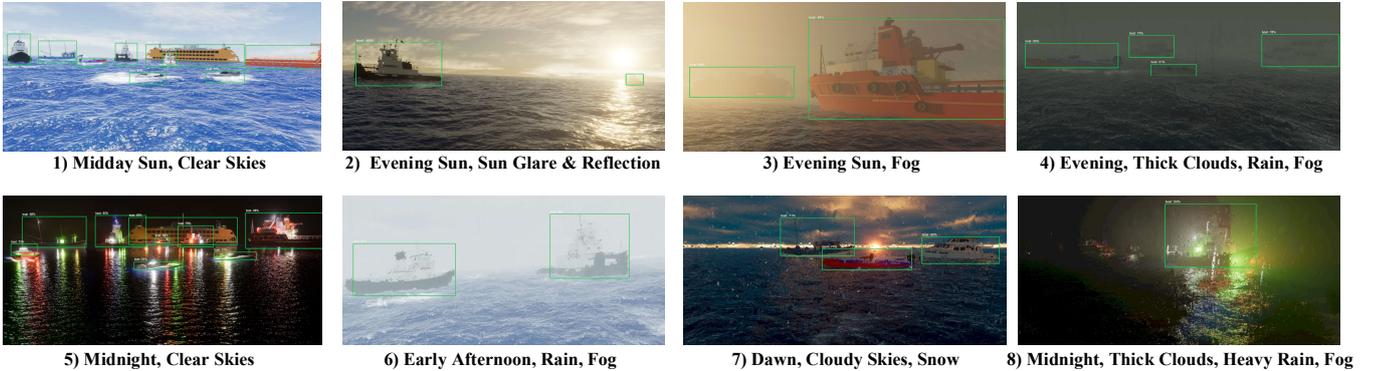

1) Midday Sun, Clear Skies   2) Evening Sun, Sun Glare & Reflection   3) Evening Sun, Fog   4) Evening, Thick Clouds, Rain, Fog

5) Midnight, Clear Skies   6) Early Afternoon, Rain, Fog   7) Dawn, Cloudy Skies, Snow   8) Midnight, Thick Clouds, Heavy Rain, Fog

**Figure 15:** Maritime Object Detection using the fine-tuned YOLOX-S model for a set of increasingly challenging cases. Zoom in for more detail.

We begin the discussion of our results with figure 12, where if we recall the training procedure in 3.A.ii., a training environment consists of other vessel Agents also being trained and some procedural terrain. At the beginning of the training process as seen in figure 12, the averaged cumulative reward is at its lowest and episode length ratio at its highest since the Agent is either experiencing collisions before an episode reset or not moving to try to reduce the penalty experienced from these collisions, from $r_S$. Due to the design of the navigational reward $r_N$, every slight positive motion towards a waypoint is rewarded and the association of this motion with the actions the Agents take is registered at around 200 episodes in addition to the need to avoid terrain. After this point, interactions with other Agent vessels start to matter more and between episodes 250 and 400, COLREG-compliance starts to become more important due to $r_D$. This is also where the episode length as a ratio of the maximum allowable episode length, $T_{max}$ starts to notably reduce from 1 as the Agent do not wander aimlessly, but plan and move towards the specified waypoint. From 450 episodes onwards, the Agents did not seem to improve significantly any further and so the training process was halted. The approximate total wall clock time of the training process was 24 hours for the rig described at the start of the Methodology section.

**COLREG-compliance for Rules 13, 14, 15**

In figure 13, we see how the Agents deal with other nearby vessels in regards to COLREG rules 13, 14 and 15. In each scenario, vessel motion trajectories are drawn with the color signifying each unique Agent. Every 10 seconds, a spherical marker is instantiated as a visual aid to the reader as this is a dynamical environment. Another aid, to provide context in the distances involved are cross-shaped gray markers which do not interact with the environment.

In the overtaking scenario we can see that, by measuring the spherical markers, there was ample spacing between the vessels, and that the slower moving vessel (yellow) waited for the faster one to pass from starboard. The faster vessel also (green) decided to navigated not in a straight line but around the slower vessel and maintained that path until nearing the waypoint at a safe distance away.

In the head-on scenario, COLREG Rule 8 in addition to 14 is observed where it can be seen that both Agents made positive actions towards avoiding collision quite early on. They keep well clear, as the distance between them peaks at approximately 80 meters until they start approaching their waypoints, well away from each other. Most importantly, it should be noted that both Agents passed by port-side, even though it would have been easier to pass by starboard-side. This action is in full accordance with Rule 14.

In the crossing scenario, Rule 15 is observed twice in succession; once at the start of the scenario and once again during a returning pass. It may be seen from the markers and from the current position of the Agent vessels, that priority in all cases was given to vessel that was on the starboard-side with the other vessel slowing down to keep out of the way.

Each one of these scenarios was repeated 100 times to test for robustness. Each time, the same outcomes as discussed above were observed.

In figure 14, we note how the Agents deal with more complex, multi-agent scenarios that incorporate terrain and vessel collision avoidance while requiring navigation to waypoints usually behind said terrain. Here the sea-state is also made slightly more unsettled with an increase in the wind-speed at 10 meters from sea-level to $6ms^{-1}$, yielding waves of approximate height of $1.25m$. In each of the images of figure 14, we observe the same COLREG behaviors shown in figure 13 being demonstrated, even in this sea-state. This is especially true in the middle case of figure 14, where four separate Agents have learnt, each by their own decision making, to follow Rules 13,14 and 15 successfully.

**Object Detection with the fine-tuned YOLOX-S Model**

Figure 15 shows bounding box predictions extracted from the fine-tuned YOLOX-S model running from within the simulation environment for various weather and visibility operating conditions. In each case, the image was the extraction of a single-frame from a recorded video from the EGO vessel's camera point of view. We note that the object detection model performs satisfactorily on some adverse operating conditions that involve fog, Sun-glare and even in cases where rain and snow has impacted the camera lens. The most challenging cases however were in environmentally darker and 'blurrier' environments such as 4, 5, 7 and especially 8, where although detections were made, they would sometimes be sporadic in nature and thus making it necessary to fall back on the phenomenological radar model instead for vessel tracking. These observations are supported by the evaluation of the model using the Average Precision metric $AP_V$ in each of these cases, as shown in figure 16 below. We also compare $AP_V$ values with the original pre-trained YOLOX-S model weights where a Curriculum Strategy was not implemented.

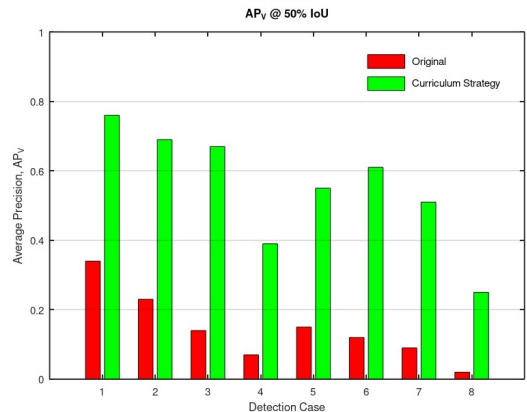

**Figure 16:** Comparison of the $AP_V$ metric at 50% $IoU$ for each weather and visibility condition for fine-tuned with Curriculum Strategy (green) and Original (red) YOLOX-S model weights.

The drop in Average Precision in the latter cases is due to the drop is recall as it becomes harder for the detector to find all the positive samples in the frame. It is clear however that the Curriculum Strategy was beneficial for the overall accuracy of the maritime object detector.

The purpose of the fine-tuning procedure, described in 3.A.ii., was to improve the overall training process of the Agents. The improvement of vessel awareness, though virtual, allows for the development of autonomous vessel behaviors that may be applied in the real-world.

**Computational Performance**

The computational performance of the entire simulation system as described in this work is summarized in figure 17. By default, the ML-Agents framework with the PPO algorithm is always used, and a performance comparison is also made when the YOLOX-S model is activated during the approach of an incoming vessel.

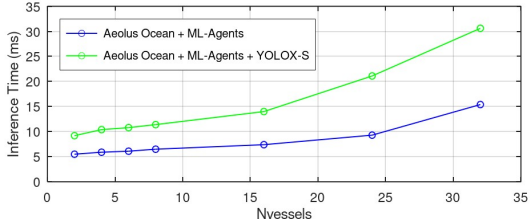

Figure 17: Computational performance of the simulation system with object detection activated (green), and without (blue) when using an NVIDIA RTX 2080Ti GPU.

It is noted in figure 17 that in all situations, the simulation runs in real-time and the frame-rate does not drop below 30 frames per second, even for the most computationally demanding scenarios.

## 5. Conclusions and Future Work

**Conclusions**

The conclusions of this work are the following:

- A realistic simulation environment for the purpose of training of DRL Agents to adopt COLREG-compliant navigational behaviors has been developed. By introducing moderate three-dimensional environmental disturbances from wind forces and oscillating waves, DRL Agents are exposed to dynamics, due to the developed physics model in this work, that are closer to reality and hence lead to smaller simulation-to-real gap. The realistic visual aspect of the simulation, which includes day/night lighting and environmental effects such as Sun-glare, Sun reflections, rain and snow, also allows for extensive testing of maritime object detectors which are used ubiquitously in autonomous vessels.

- A DRL Agent can learn to generalize, to a certain extent at a satisfactory level, its COLREG-compliant navigation in unseen scenarios after having experienced through trial-and-error the mistakes of its actions in controlled training environment. In its navigation to a specified waypoint the DRL Agent, while experiencing full scale of rigid-body motion, shows it can avoid collisions with other vessels, terrain (including underwater terrain such as reefs) produced by the procedural methods outlined in this work, while dealing with moderate external environmental disturbances such as wind and waves.

- A set of novel reward functions have been developed that lead to the creation of the aforementioned COLREG-compliant behaviors. A significant effort was made to reduce the reliance of additional hyper-parameters required for these reward functions. The concept of a time-safety parameter was also introduced to provide a clearer trade-off point between navigational and collision-specific rewards.

- A phenomenological sensor-fusing approach has been implemented in this work that combines a simulated suite of sensors common to marine vessels (sonar, marine radar, visual). While it is acknowledged that the real-world operation of these sensors is different, the focus in this work was their outputs, namely the plausible tracking data that allowed the DRL Agents to observe and make decisions in their environments. For the visual sensor, a Curriculum Learning Strategy was adopted for the effective fine-tuning of the YOLOX-S object detection model weights, leading to satisfactory performance in challenging weather and visibility conditions.

**Future Work**

There are also some weakness that were exposed in regards to the approach detailed in this work, and merit further investigation for ways in which they could be dealt with, or at least mitigated.

- The way in which the reward functions have been set up essentially assign a single final scalar reward to the DRL Agent, summed over all competing sub-rewards (navigation, collision, time-penalty). This however might not be enough to deal with the complexity of all potential scenarios. In a rare example, the temporary distancing from a waypoint under certain conditions might lead to a larger negative reward than the collision avoidance with another vessel, hence leading to a dangerous situation. While this may be mitigated with further calibration and redevelopment of the reward functions, a more fruitful approach would be to have not one large general actor-critic network, but multiple smaller ones, with the output of one network (related to a collision avoidance) feeding in as an observational input to the navigational-related network. It remains to be seen whether this sequential decision making way is more effective in more complex scenarios, but is an interesting avenue of investigation.

- The extent by which weather and visibility conditions affect the COLREG-compliant navigation of the DRL Agent is another interesting avenue for exploration. Large waves from stormy sea-states at different times of the day introduce more complex dynamics and it is uncertain how far the current approach is able to with these disturbances effectively.

- Further development in more realistic sensors, including the aspect of fusing their data together, in such a simulation environment, would find more traction in real-world applications as the inputs and outputs would more closely relate and thus aid in further reducing the simulation-to-real gap.

- The ambiguity in interpreting COLREGs [61][62] makes it difficult to ascertain quantifiably to what extent the navigation of a DRL Agent satisfied a certain rule. Ensuring ways in which COLREGS could be interpreted via some measurable value would greatly aid in the development of autonomous vessel technologies.

We hope that the work outlined in this paper leads to potential avenues for developments and improvements in the field of autonomous marine vessel navigational guidance and encourages the usage of the three-dimensional simulation systems towards this goal.

**Acknowledgement**

The author would like to acknowledge the generous funding and support of the Stavros Niarchos Foundation (SNF) Industrial Fellowship Program. Thanks are due to Dr. Stavros Perantonis, Director of the Computational Intelligence Laboratory, for the generous provision of computational resources that enabled this work.